%% file: main.tex
\documentclass[10pt]{article}
\usepackage[preprint]{tmlr}
\usepackage{booktabs}
\usepackage{graphicx}
\usepackage{amsmath,amssymb}
\usepackage{microtype}
\usepackage[hidelinks]{hyperref}
\usepackage{url}

\title{Measurement Boundaries in LLM Financial Agent Evaluation:\\
Fixed-Tape Execution and Multi-Defect Auditing}

\author{Weicheng Xue\\Virginia Tech\\\texttt{weich97@vt.edu}}
\hypersetup{pdfauthor={Weicheng Xue},
  pdftitle={Measurement Boundaries in LLM Financial Agent Evaluation: Fixed-Tape Execution and Multi-Defect Auditing}}

\begin{document}
\maketitle

\begin{abstract}
What controls are needed to interpret execution performance and audit scores in
LLM agent evaluations? We study two limits on these interpretations in a
financial agent harness.
In Study~A, comparing independent runs under idealized and stressed execution on
three synthetic settings that share one 24-day upward phase mixes
the execution rule with fresh model responses and portfolio feedback: the parsed
decision paths agree in only $19.8\%$ of $450$ pairs. Replaying each stored
response tape through both execution destinations gives a narrower result.
Conditional on those responses, stressed execution changes total return by
$-0.0170$ (95\% interval $[-0.0230,-0.0117]$), or $10.4\%$ of the idealized
baseline, and ten seed clusters do not resolve the model ranking.
Study~B corrects an incomplete answer key and replaces legacy tasks with
matched zero-, one-, and two-defect tasks under an explicit multi-label prompt.
The drop in target violation recall from
one to two defects is positive in five of six combinations of auditor and source (median
$0.267$), with three surviving Holm correction. Yet the auditor that includes
both target labels most often has micro-precision $0.149$, emits findings on
$98/100$ zero-defect tasks, and returns the exact dual-defect set in only
$21/100$ cases. Target recall by itself therefore gives a poor account of audit
quality on this construction. The studies address different limits: what an
execution comparison estimates, and what target recall captures. Together, they
show how fixed conditions and diagnostic controls bound the claims a score can
support.
\end{abstract}

\section{Introduction}\label{sec:intro}

An LLM agent can be evaluated by the outcomes of its actions, and an LLM auditor
by the defects it finds in an action record. Neither score explains its own
scope. A simulator turns intended orders into fills, and a scorer decides which
findings count. We ask: \emph{what controls are needed for execution performance
and audit scores to support the claims made from them?} We study two distinct
limits on those claims in a financial agent harness.

Study~A begins with a common but ambiguous comparison: run the same model under
idealized and stressed execution and compare the outcomes. The two runs do not
differ only in execution. They draw fresh provider responses, and once their
trades differ they also expose the model to different portfolio states. We use a
stored-response replay to answer a narrower question: what does the execution
mechanism change when the provider response sequence is held fixed? The replay
yields a measurable contrast on our synthetic matrix, and ten seed clusters do
not resolve whether the top order is preserved.

Study~B concerns a scoring failure. Our original answer key listed one defect
where the generator had made two stated rules false. The legacy prompts also
contained quantity cues, phrased differently across domains. Re-scoring credits
valid edit findings but neither recovers an unreported target violation nor
identifies why the auditor omitted it. We therefore collected
matched zero-, one-, and two-defect tasks under a prompt that permits multiple
findings. Target violation recall falls in five of six cells, but the auditors'
specificity is poor and differs sharply. deepseek-v4-pro includes both target
labels more often, yet produces the exact dual-defect set in only $21/100$ cases
and emits findings on $98/100$ zero-defect tasks.

These cases address two parts of agent evaluation: measuring action outcomes and
measuring the quality of checks on action records. Study~A asks whether a
comparison estimates the intended effect; Study~B asks whether the scored
success captures the intended audit quality.

\paragraph{Contributions.}
\begin{enumerate}
  \item An empirical distinction between independently sampled closed-loop
  comparisons and fixed-response execution contrasts. Parsed decision paths
  agree in $19.8\%$ of $450$ source pairs, showing that the source comparison
  does not hold the action sequence fixed. A two-origin, two-destination replay
  estimates a conditional mechanical contrast and quantifies its uncertainty.
  \item A two-model, matched multi-defect audit experiment whose target recall,
  false-positive, precision, and exact-set outcomes are reported together. The
  plan and corpus are internally hash-bound.
\end{enumerate}

\section{Related Work}\label{sec:related}

\paragraph{LLM agents for trading and financial decision making.}
A growing literature equips language models with memory, tools, and role
structure, and then reports trading performance. \citet{yu2023finmem} add a
layered memory and a character profile; \citet{yu2024fincon} introduce a
hierarchy of managers and analysts with a risk-control component and conceptual verbal
reinforcement; \citet{xiao2024tradingagents} use debates between bullish and bearish
researchers and a risk-management team in a multi-agent trading framework. Domain foundation
models
\citep{wu2023bloomberggpt,yang2023fingpt} and environment libraries such as
FinRL-Meta \citep{liu2022finrlmeta} supply the substrate, and a recent survey
\citep{ding2024tradingsurvey} reviews backtesting-based evaluation of trading
agents. These works motivate studying how measured performance depends on both
agent design and evaluation choices. Study~A focuses on the execution
convention (fill price, fill timing, and cost attribution) by holding one
realized response tape fixed and measuring the downstream mechanical contrast
between two destinations.

\paragraph{Execution realism, microstructure, and backtest overfitting.}
The cost of trading is not a nuisance term. Adverse selection produces a spread
even with risk-neutral, zero-profit market making \citep{glosten1985}, and price
impact is linear in order flow in the canonical strategic-trading model
\citep{kyle1985}; \citet{almgren2001} formalise the resulting
trade-off between price impact and risk in optimal execution, and \citet{hasbrouck2007}
gives the econometric treatment. Empirically, \citet{frazzini2018trading}
measure realised costs on \$1.7 trillion of live institutional executions and
find them an order of magnitude below common academic estimates, so
misspecifying the convention can move results in either direction. In
parallel, repeated strategy search inflates backtest performance
\citep{bailey2014pseudo,lopezdeprado2018}; selection-adjusted Sharpe inference
addresses one consequence of that search, as a deflated Sharpe ratio
\citep{bailey2014deflated} or a multiple-testing haircut
\citep{harvey2015backtesting}; and related multiple-testing concerns arise in
factor discovery \citep{harvey2016cross}.
This body of work models the cost function as an object to be estimated, or the
search process as an object to be corrected. This report treats the destination
as an experimental factor in a controlled replay, a record-and-replay argument
in the sense of \citet{leblanc1987replay}. Replay is the right tool when the
intended estimand holds realized responses fixed, particularly because
nominally deterministic LLM settings need not reproduce identical outputs
\citep{atil2024nondeterminism}.

\paragraph{LLM-as-a-judge and its biases.}
\citet{zheng2023judge} established strong LLMs as scalable proxies for human
preference and catalogued position, verbosity, and self-enhancement biases;
\citet{wang2024fair} show that pairwise rankings can be flipped by reordering
candidates alone, \citet{panickssery2024self} link self-preference to
self-recognition, and \citet{gu2024judgesurvey} survey the resulting mitigation
literature. Closer to defect detection, \citet{ullah2024vulnerabilities} and
\citet{ding2025primevul} show that LLM security auditors are non-robust and
that prevailing benchmarks overstate their capability. Study~B identifies a
different measurement problem: a corpus containing only single-defect instances
cannot estimate behaviour on co-occurring defects. Our new experiment adds the
missing conditions and shows why target recall must be reported with
specificity.

\paragraph{Scalable oversight.}
Debate \citep{irving2018debate}, iterated amplification
\citep{christiano2018amplification}, recursive reward modelling
\citep{leike2018reward}, and assisted self-critique
\citep{saunders2022critique} propose ways to supervise systems that humans
cannot directly evaluate, building on learning from human preferences
\citep{christiano2017preferences}; \citet{bowman2022oversight} make the claim
empirically testable via sandwiching, \citet{amodei2016concrete} frame the accident-risk
setting, and \citet{greenblatt2024control} evaluate oversight protocols against
an intentionally subversive policy. These works propose and measure oversight
protocols. Study~B asks whether the benchmark used to evaluate an auditor can
itself shape the conclusion, and replaces a single-defect construction with
matched zero-, one-, and two-defect tasks.

\paragraph{Benchmark and evaluation fragility.}
LLM scores move under meaning-preserving perturbations of prompt formatting
\citep{sclar2024format}, instruction paraphrase \citep{mizrahi2024multiprompt},
and adversarial prompt edits \citep{zhu2023promptrobust}; leaderboard rankings
shift by up to eight positions under trivial changes to answer ordering or
extraction \citep{alzahrani2024leaderboards}. \citet{liang2023helm} argue for
multi-metric, multi-scenario reporting. \citet{kapoor2024agents} show that
evaluation focused on accuracy can overlook cost and lead to mistaken
conclusions about the benefits of agent complexity.
\citet{yao2024taubench} show that agent success is inconsistent across repeated
trials, with pass$^8$ below $25\%$ in their retail domain.
This report extends fragility from surface form to two structural conventions.
The execution convention has been varied as an ablation before, including in the
substrate paper \citep{xue2026tradearena}, and recorded-decision replay is an
established device rather than a contribution of this report. Study~A crosses
both response origins with both execution destinations on a five-identifier,
24-period matrix and examines conditional contrasts and unresolved ranking
uncertainty. Study~B uses separate audit-task collections and examines what
target recall does and does not measure.

\paragraph{Fault injection and mutation testing.}
Measuring a checker by seeding faults is old: the coupling effect motivates
mutation adequacy \citep{demillo1978hints}, \citet{jia2011mutation} survey the
field, and \citet{just2014mutants} find that mutant detection is correlated with
real fault detection in the programs they studied. Analogues in hardware and
systems run from classical fault injection
\citep{hsueh1997faultinjection} to chaos engineering \citep{basiri2016chaos}.
This report applies that discipline to an LLM auditor, where matched single-
and multi-defect conditions expose differences in target violation recall.

\section{Two Boundaries Between Scores and Claims}\label{sec:framework}

The studies follow the same reasoning: identify the claim, state what the
original comparison or score measures, add controls for what it leaves open,
and limit the conclusion to the resulting evidence. It organizes the two
studies; it is not a claim that they share one source of error.

\paragraph{The comparison boundary.}
In Study~A, an independently sampled closed-loop comparison allows responses and
portfolio states to differ. Feedback is part of a total execution effect, not
automatically a nuisance; deterministic policies can also react to changed
state. A fixed-response replay blocks that feedback and measures a narrower
mechanical contrast. It does not correct the closed-loop comparison into an
estimate of the same quantity. Path disagreement shows that responses were not
held fixed, not that the comparison is biased. We treat the observed diagonal
as descriptive because the collection does not separately identify
provider-time changes, sampling, and adaptation.

\paragraph{The score-coverage boundary.}
In Study~B, target recall measures whether a specified violation is reported.
Even a correctly computed recall score does not measure unsupported findings
or complete audit correctness. Matched source cases with zero, one, and two
defects test behaviour beyond a single-defect condition. Complete rule-based
keys, zero-defect controls, precision, and exact-set accuracy then separate
target coverage from correct auditing.

\section{Shared Substrate and Protocol}\label{sec:substrate}

Both studies use one codebase introduced in earlier work
\citep{xue2026tradearena}. It records signal, intended target weight,
risk-gate revision, order conversion, simulated execution, and portfolio update
as serialized artifacts. Study~A replays the execution stage; Study~B audits
injected trajectory defects.

\paragraph{Models.} Studies~A and~B record the direct provider and API model
identifier used for each response. An identifier without a dated revision is a
provider snapshot, not an immutable model version. Study~A uses five identifiers
(two DeepSeek and three GLM); Study~B uses deepseek-v4-pro and glm-5.

\paragraph{Provenance.} Runs carry manifests and content digests, and the
released scripts compare derived tables against recorded values. These are
consistency checks; the Reproducibility statement gives their exact scope.

\paragraph{Replay.} Provider responses are cached with per-task checkpointing,
so the reported analyses reuse stored responses. The multi-label collection
contains $600$ provider responses, and the corrected self-consistency arm contains
$720$ responses at temperature $0.7$. The fixed-tape replay and answer-key
re-scoring run locally without new provider calls.

\section{Study A: A Fixed-Tape Contrast for Execution Mechanics}\label{sec:studyA}

Independent E0 and E1 runs change the execution rule and, through fresh
sampling and portfolio feedback, the realized response path as well. We
therefore replay each stored response tape through both execution destinations.
The replay isolates the mechanical execution contrast conditional on fixed
responses, and it shows how little ten seed clusters say about the model
ranking.

\subsection{What independent runs compare}

The completed direct-API matrix contains five recorded model identifiers,
three synthetic market settings, two execution levels (E0, an immediate
full-fill reference simulator; E1, a fixed stress simulator), ten seeds, and
three provider samples: $900$ runs.
The market settings vary volatility, trend scale, and tail behaviour, but they
all cover the same 24-day upward phase; across the fixed seeds and two assets,
every generated price path ends above its starting value. E1 uses one-step
latency and a $5\%$ participation cap. Both arms charge one basis point of
commission and start from two basis points of slippage; E1 additionally applies
$0.15$ times realized participation and $0.1$ times intraday bar volatility to
its slippage rate.

The diagonal comparison pairs an independently sampled E0 run with an
independently sampled E1 run. Across $450$ such pairs, which share their initial
prompt, the first raw response agrees only $44.9\%$ of the time and the parsed
24-period decision path agrees in $19.8\%$. Of the pairs whose parsed responses
eventually differ, $61\%$ already differ at the first response, before execution
can feed anything back. The remaining paths may also diverge through portfolio
feedback. The diagonal therefore combines a change in execution with fresh
sampling, provider time, and a legitimate closed-loop response to different
state; it is not an execution-only contrast.

\subsection{The fixed-response control and its estimand}

For each stored provider-response tape, we replay the recorded responses through
both execution destinations. The replayer returns the stored response in sequence
even when the portfolio-dependent prompt generated at the other destination
would differ. It therefore fixes the pre-risk decision path by construction.
Risk approval and fills remain downstream, but the route from execution through
the next portfolio prompt back into the model is blocked.

Let $Y_{od}$ be an outcome of the tape recorded under origin
$o\in\{\mathrm{E0},\mathrm{E1}\}$ and replayed at destination
$d\in\{\mathrm{E0},\mathrm{E1}\}$, averaged over the $450$ matched pairs. The
within-tape destination contrast is $D_o = Y_{o,\mathrm{E1}}-Y_{o,\mathrm{E0}}$,
and the estimand is the \emph{balanced destination} contrast
$\bar D=(D_{\mathrm{E0}}+D_{\mathrm{E1}})/2$: what the stress simulator changes
for the recorded decisions. The \emph{balanced tape-origin} contrast $\bar O$
averages $O_d = Y_{\mathrm{E1},d}-Y_{\mathrm{E0},d}$ over $d$ and is descriptive,
because origin bundles sampling, feedback, collection time, and provider state.
The two sum to the observed diagonal
$Y_{\mathrm{E1},\mathrm{E1}}-Y_{\mathrm{E0},\mathrm{E0}}$, and the interaction
$D_{\mathrm{E1}}-D_{\mathrm{E0}}$ measures how much the destination effect
depends on which tape is replayed.

\subsection{What the fixed-response control supports}

\begin{figure}[t]
\centering
\includegraphics[width=\linewidth]{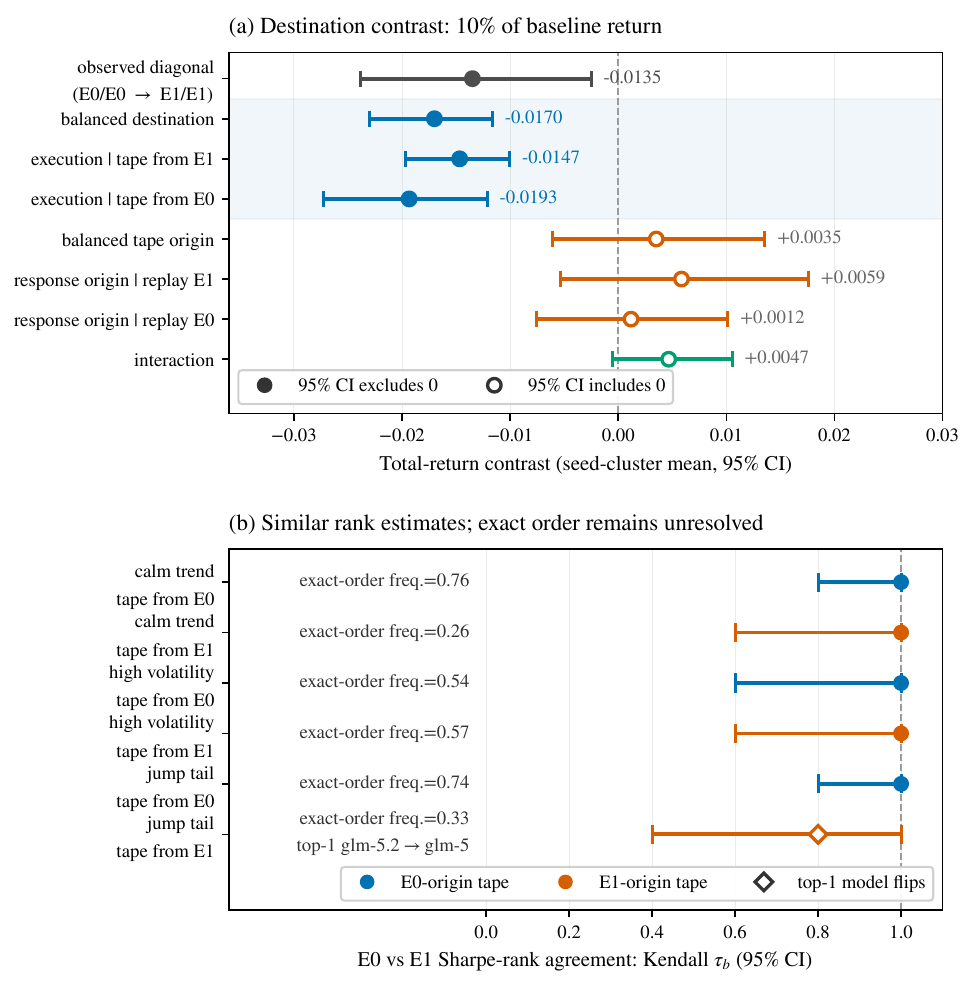}
\caption{The fixed-tape execution contrast. (a) The $2\times2$ decomposition on
total return. The downstream mechanical contrast excludes zero under either tape
origin; the tape-origin term is descriptive rather than causal. (b) Kendall
$\tau_b$ between E0 and E1 Sharpe rankings under a fixed tape, with the frequency
of exact-order agreement across cluster-bootstrap resamples. Confidence
intervals come from a bootstrap over the $10$ shared seed clusters;
rank-correlation values are discrete because only five models are ranked.}
\label{fig:execid}
\end{figure}

Figure~\ref{fig:execid} reports the decomposition. On total return, $\bar D$ is
$-0.0170$ with a seed-cluster bootstrap interval of $[-0.0230,-0.0117]$, equal
to $10.4\%$ of the $+0.164$ E0 baseline, and $\bar O$ is $+0.0035$
($[-0.0061,+0.0136]$). The observed diagonal, $-0.0135$, is their sum. A reader
who takes the diagonal as the execution effect attributes part of a sampling
and feedback difference to the simulator; a reader who takes $\bar D$ as the
total effect ignores that an adaptive agent can answer E1 fills with different
later decisions.

The split is metric-specific. On Sharpe, $\bar O$ is $+0.813$
($[+0.441,+1.239]$), and $\bar D$ is $21.3\%$ of baseline Sharpe, compared with
$10.4\%$ of baseline return.

\paragraph{Uncertainty with ten seeds.}
A $t_9$ interval based on the cluster standard error is
$[-0.0239,-0.0101]$ for the fixed-tape total-return contrast, compared with the
bootstrap interval $[-0.0230,-0.0117]$. Recomputing after dropping each seed in
turn moves the point estimate within $[-0.0181,-0.0152]$. This leave-one-seed
range is a sensitivity diagnostic, not an uncertainty interval. The same
distinction matters for the total-return tape-origin term: its leave-one-seed
range is $[+0.0005,+0.0061]$, while its bootstrap interval includes zero.

\paragraph{Ranking uncertainty and activity.}

Under a fixed tape the E0 and E1 Sharpe rankings have $\tau_b = 1.000$ in five
of six combinations of setting and tape origin and $+0.800$ in the sixth
(Table~\ref{tab:ranking}). These point estimates do not establish stability. The
frequency of exact-order agreement across seed-cluster bootstrap resamples ranges
from $0.26$ to $0.76$, and the top-ranked model changes from glm-5.2 to glm-5 in
the jump--tail setting with tapes from E1. At ten seed clusters, the ranking question is
unresolved. The apparent agreement is also helped by an almost inactive row:
deepseek-v4-pro records zero return, all-hold, and zero gross exposure in $358$
of $360$ replay rows, so it is last in nearly every ranking and contributes
several concordant pairs mechanically.

\begin{table}[t]
\centering
\caption{Sharpe-rank comparison under a fixed response tape. Kendall $\tau_b$
is near one, but exact-order agreement occurs in only $0.26$--$0.76$ of
seed-cluster bootstrap resamples, and the winner changes in the jump--tail
setting with tapes from E1. ``Exact order'' is a bootstrap agreement frequency, not a posterior
probability that the population order is correct.}
\label{tab:ranking}
\small
\begin{tabular}{llrlrl}
\toprule
Setting & Tape & $\tau_b$ & 95\% CI & Exact-order freq. & Winner E0 $\to$ E1 \\
\midrule
calm trend      & E0 & 1.000 & [0.80, 1.00] & 0.7635 & glm-5.2 $\to$ glm-5.2 \\
calm trend      & E1 & 1.000 & [0.60, 1.00] & 0.2608 & glm-5.2 $\to$ glm-5.2 \\
high volatility & E0 & 1.000 & [0.60, 1.00] & 0.5368 & glm-5.2 $\to$ glm-5.2 \\
high volatility & E1 & 1.000 & [0.60, 1.00] & 0.5678 & glm-5.2 $\to$ glm-5.2 \\
jump--tail      & E0 & 1.000 & [0.80, 1.00] & 0.7369 & glm-5.2 $\to$ glm-5.2 \\
jump--tail      & E1 & 0.800 & [0.40, 1.00] & 0.3265 & glm-5.2 $\to$ \textbf{glm-5} \\
\bottomrule
\end{tabular}
\end{table}

The inactive row is a substantive warning. A model that almost never trades is
largely insensitive to execution, has little drawdown, and pays little slippage.
A leaderboard should report activity beside robustness so that inaction is not
mistaken for successful adaptation (Figure~\ref{fig:pooling}a).

\paragraph{Supported claim.}
For the stored responses on this 24-day upward synthetic matrix, replacing the
immediate reference fills with the specified stress simulator lowers total
return by $0.0170$ on average; the model ranking under either simulator remains
unresolved at ten seed clusters.

\section{Study B: What a Single-Defect Audit Score Misses}\label{sec:studyB}

Target recall records whether one specified violation is mentioned. We first
correct the legacy answer key, then collect matched zero-, one-, and two-defect
tasks and report target recall beside false positives, precision, and exact-set
accuracy.

\subsection{What target recall measures}

We take a valid agent trajectory, inject defects by a known procedure, and score
the auditor's structured findings against the resulting rule violations.
Injection alone does not guarantee an exhaustive answer key, as the legacy error
below shows. For the new corpus, the generator and analyzer contain separately
implemented rule oracles over the four stated rules in each domain. A task is
admitted only when the complete set of violated (step, rule) pairs exactly matches
its zero-, one-, or two-defect key. This establishes ground truth relative to
those coded rules, not for defects outside them.
Defects span a difficulty ladder by construction: a
single-record constraint check (L1), cross-record reconciliation (L2), and
recomputation (L3). Auditors are two directly served API snapshots,
deepseek-v4-pro and glm-5, run under a neutral audit-role prompt at temperature~0.
Target recall credits the specified violation when it appears in a response.

\paragraph{A defect in our own answer key.}

Our first version of this benchmark interpreted lower recall of constraint
violations as a preference for an alternative reading of the same step. That
interpretation does not survive scrutiny of how the tasks were built: the
alternative label describes another real defect, not a benign explanation.

The confounded variant differs from the one-defect variant in a single field: the recorded
intent for the defective symbol. Nothing is deleted to build it. The injected
approval already carries no intervention record in either arm. Moving the intent
away from the approved value is therefore enough to make a second stated rule
\emph{actually} false, because the auditor is told that an approval differing
from its intent must record the intervention, and none does. Two of the stated
rules are violated at that step. The answer key listed one. An auditor that
reported the unrecorded edit was therefore assigned a false-positive penalty
for a valid finding. Correcting the key removes that penalty and changes
set-level scoring; it does not turn an unreported constraint violation into a
hit, so the old score mixed a genuine omission with an incorrect penalty for a
valid finding.

\paragraph{Quantity cues in the legacy prompts.}

\begin{figure}[t]
\centering
\includegraphics[width=\linewidth]{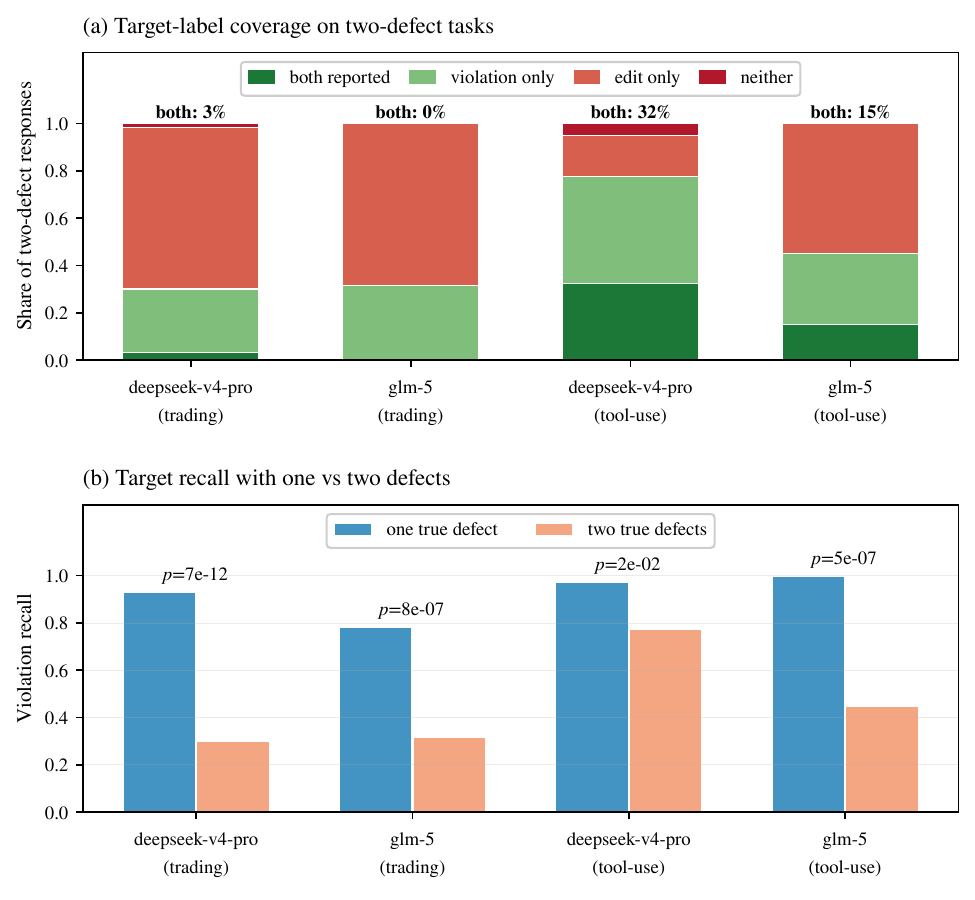}
\caption{Behaviour on the legacy corpus, whose prompts contain domain-specific
quantity cues. (a) On tasks where two defects are objectively true, the share
of responses reporting both target labels, one, or neither; this is not the
total number of findings in a response. (b) Violation recall with one true
defect versus two, with within-pair McNemar $p$-values, scored against the
corrected key so that reporting the edit is credited rather than penalized. The
low frequency of ``both'' does not identify a quantity-cue effect.
The new matched experiment uses an explicit multi-label instruction.}
\label{fig:cardinality}
\end{figure}

Re-scoring the stored responses against a key in which the confounded arm carries
two true defects credits valid edit findings and removes the incorrect precision
penalty. It also shows what the auditors actually reported. On the two-defect arm
they report \emph{exactly one} of the two true defects in $62$--$100\%$ of tasks
and both in $0$--$33\%$; glm-5 never reported both in any of its sixty trading
pairs (Figure~\ref{fig:cardinality}).

All $120$ trading prompts state that exactly one record was tampered with;
all $80$ tool-use prompts state ``Exactly one step contains a single injected
defect.'' These are different quantity cues. One tampered record can violate
multiple rules, so the trading instruction does not logically require one
label. Moreover, reporting one of the two true labels is not the same as
reporting only one finding in the entire response. In $4/60$ of deepseek-v4-pro's
legacy trading dual-defect tasks, exactly one target label is reported but the
response contains multiple findings. The legacy design cannot distinguish
quantity-cue effects from other causes of omission; it does not establish that
one-label behaviour is instruction compliance.

The drop in target violation recall remains after correcting the key, because a valid
edit finding does not substitute for the target violation (deepseek-v4-pro
$0.93\!\to\!0.30$ and glm-5 $0.78\!\to\!0.32$ in trading; $0.97\!\to\!0.78$ and
$1.00\!\to\!0.45$ in tool-use; within-pair McNemar $p = 7\times10^{-12}$,
$8\times10^{-7}$, $2\times10^{-2}$, $5\times10^{-7}$). We therefore use a new
matched multi-defect design with an explicit multiple-finding instruction for
the primary analysis.

\subsection{Matched controls and the primary estimand}

Re-scoring stored responses is a corrective analysis of data collected for
another purpose. We therefore collected responses for matched triplets of the
same source case carrying zero, one, and two true defects: $300$ tasks over
$100$ triplets, two auditors, temperature~0, and one sample per task. The source
cases come from three collections, named consistently in the tables and figures:
\emph{trading A}, valid trading trajectories produced by deepseek-v4-pro;
\emph{trading B}, valid trading trajectories produced by glm-5; and
\emph{tool-use}, rule-generated tool-call logs of a synthetic operations agent.

One trading triplet shows the construction. The source step approves a target
weight for one symbol under a stated per-name cap of $0.35$. In the zero-defect
variant the approval keeps the value the risk gate actually produced and, where
that value differs from the model's intent, carries the required clip record.
In the one-defect variant the intended and approved weights are both set to
$0.56$ and the clip record is removed, so exactly one stated rule is false: an
approval above the cap without clip evidence. The two-defect variant changes a
single field, the intended weight, to $0.28$; the approval now also differs
from the intent without an intervention record, so a second stated rule is
false at the same step. A complete audit therefore returns the empty set,
$\{\text{unclipped position}\}$, and $\{\text{unclipped position},
\text{silent risk edit}\}$ respectively, and target recall asks only whether
the first label is present. The tool-use triplets mirror this with a per-tool
argument cap and a silent change to an approved argument.

The primary
estimand is $\Delta_c=R_{\text{single}}-R_{\text{dual}}$ within matched triplets,
tested by two-sided exact McNemar tests with Holm correction across six
combinations of auditor and source.

The task tree contains an analysis plan and manifest linked by SHA-256 digests,
and the completed grid contains $600/600$ unique (auditor, task, sample) keys. The
digests establish agreement among the released plan, corpus, and result grid,
not temporal precedence, because all three remain in an author-writable tree
without a third-party timestamp; we therefore call the design internally
hash-bound rather than preregistered.

This corpus does \emph{not} assert cardinality: its prompt says the
artifact ``may contain zero, one, or multiple defects'' and asks for one finding
per violated rule. That makes the answer counts and specificity outcomes
recorded in the plan interpretable (Tables~\ref{tab:cardinality} and
\ref{tab:audit-specificity}).

\subsection{What the matched controls support}

Table~\ref{tab:frozen} reports the primary outcome. The drop is positive in five
of six cells, its median is $0.267$, and three cells survive Holm correction.
Nine of $600$ responses fail to parse ($1.5\%$); malformed output is never
credited as a correct empty answer.

Once the prompt permits multiple findings, the two auditors differ sharply in
whether the two target labels appear. deepseek-v4-pro includes both in
$70$--$88\%$ of dual-defect responses across the three sources, whereas glm-5
does so in $0$--$48\%$ (Table~\ref{tab:cardinality}). Performance on tasks
containing only one defect therefore does not determine behaviour on this
two-defect construction.

\begin{table}[t]
\centering
\caption{Presence of the two target labels on the two-defect arm. The four
columns partition responses only by whether the target violation and target edit
appear. Additional or duplicate findings do not change a response's column, so
``both targets'' is not exact-set accuracy and ``violation target'' does not mean
that no unsupported finding was emitted.}
\label{tab:cardinality}
\small
\begin{tabular}{llrrrr}
\toprule
Auditor & Source & Both targets & Violation target & Edit target & Neither target \\
\midrule
deepseek-v4-pro & tool-use & \textbf{35} & 3 & 0 & 2 \\
deepseek-v4-pro & trading A & \textbf{22} & 0 & 7 & 1 \\
deepseek-v4-pro & trading B & \textbf{21} & 0 & 9 & 0 \\
glm-5 & tool-use & 19 & 0 & 21 & 0 \\
glm-5 & trading A & 4 & 9 & 16 & 1 \\
glm-5 & trading B & 0 & 5 & 25 & 0 \\
\bottomrule
\end{tabular}
\end{table}

\begin{table}[t]
\centering
\caption{Target-label counts beside specificity outcomes. ``Zero any'' is the
number of zero-defect tasks with at least one emitted finding. ``Zero exact'' is
a parse-valid, duplicate-free empty finding set; parse failures are not credited.
``Dual both'' requires both target labels but permits extras, whereas ``dual
exact'' requires a parse-valid, duplicate-free match. Micro-precision pools all
findings over the $300$ tasks per auditor, and overall exact set applies the same
strict match to all $300$.}
\label{tab:audit-specificity}
\small
\begin{tabular}{lrrrrrr}
\toprule
Auditor & Zero any & Zero exact & Micro-P & Exact/300 & Dual both & Dual exact \\
\midrule
deepseek-v4-pro & $98/100$ & $1/100$  & 0.149 & $29/300$  & $78/100$ & $21/100$ \\
glm-5           & $45/100$ & $55/100$ & 0.199 & $108/300$ & $23/100$ & $7/100$ \\
\bottomrule
\end{tabular}
\end{table}

Despite including both targets more often, deepseek-v4-pro has lower overall
exact-set accuracy ($29/300$) than glm-5 ($108/300$), because it also emits
unsupported findings; on the dual-defect tasks alone its exact-set accuracy is
nevertheless higher ($21/100$ against $7/100$). Neither model is well described
by a single ``completeness'' score.

The new and re-scored estimates also differ in magnitude and, for
deepseek-v4-pro on tool-use tasks, in sign. The datasets differ in task construction and
cardinality instructions, so the legacy values are prompt-specific retrospective
contrasts, not a replication or an upper bound. We use the new collection for the
primary claim and retain the legacy analysis to document the answer-key and
prompt defects.

\paragraph{Supported claim.}
On this matched violation-plus-edit construction, recall measured with one
defect does not carry over to the same case with two, and coverage of both
targets is a different quantity from exact correctness. The remaining analyses
use legacy tasks to examine prompt sensitivity and self-audit and are secondary.

\begin{table}[t]
\centering
\caption{Internally hash-bound multi-label experiment. Target violation recall
on the one-defect and two-defect conditions within matched triplets, with exact
two-sided McNemar tests and Holm correction across six cells. Positive drop means
that the target violation label appears less often when the specified edit defect
also occurs.}
\label{tab:frozen}
\small
\begin{tabular}{llrrrrr}
\toprule
Auditor & Source & $n$ & Single & Dual & Drop & Holm $p$ \\
\midrule
deepseek-v4-pro & tool-use & 40 & 0.900 & 0.950 & $-0.050$ & 0.625 \\
deepseek-v4-pro & trading A & 30 & 0.900 & 0.733 & $+0.167$ & 0.125 \\
deepseek-v4-pro & trading B & 30 & 0.967 & 0.700 & $+0.267$ & \textbf{0.031} \\
glm-5 & tool-use & 40 & 1.000 & 0.475 & $+0.525$ & $\mathbf{6\times10^{-6}}$ \\
glm-5 & trading A & 30 & 0.833 & 0.433 & $+0.400$ & \textbf{0.009} \\
glm-5 & trading B & 30 & 0.433 & 0.167 & $+0.267$ & 0.116 \\
\bottomrule
\end{tabular}
\end{table}

\subsection{Prompt sensitivity of target violation recall}

We next ask whether three changes to prompts and sampling alter the presence of the
target violation label on the legacy two-defect tasks. These analyses measure
targeted recall, not exact-set accuracy or overall audit quality. We re-audit the
same matched pairs under an explicit \emph{constraint-check} instruction that
names the rule, a generic \emph{step-by-step} instruction, and
\emph{self-consistency} using three samples at temperature $0.7$ with majority
aggregation. Each arm contains $n=60$ matched pairs per auditor, pooled over the
two producers. The six arm-by-auditor comparisons are exploratory; the McNemar
$p$-values in this subsection are unadjusted.

Naming the rule produces the largest increase in target violation recall. Violation recall rises
from $0.300$ to $0.833$ for deepseek-v4-pro ($34$ pairs fixed against $2$
regressed, McNemar $p = 2\times10^{-8}$) and from $0.317$ to $0.933$ for glm-5
($38$ fixed, $1$ regressed, $p = 1\times10^{-10}$), while recall on the legacy
one-defect arm stays
high ($0.850$ and $0.883$). Precision and exact-set outcomes are not reported
for these arms, so the gain concerns targeted recall only.

Generic deliberation changes targeted violation recall differently across the
two auditors: deepseek-v4-pro moves from $0.300$ to $0.333$ ($p = 0.73$), while
glm-5 moves from $0.317$ to $0.617$ ($24$ fixed against $6$ regressed,
$p = 0.001$).

Self-consistency splits the auditors the other way. Using a majority of three
samples at temperature $0.7$ increases recall for deepseek-v4-pro from $0.300$
to $0.500$ ($16$ fixed against $4$ regressed,
$p = 0.012$). For glm-5, observed recall decreases from $0.317$ to $0.283$,
but the difference is not statistically detectable ($p = 0.77$).
An \emph{any-of-three} rule would yield $0.883$ and $0.433$, but it
credits a finding from any single sample and so inflates recall mechanically;
we show it only to indicate how much the aggregator matters. The arm also varies
temperature alongside sampling, so it is not a single-factor contrast.

For these two auditors on this legacy task set, naming the constraint raises
observed targeted recall for both; generic deliberation does so only for glm-5,
and majority self-consistency only for deepseek-v4-pro.

Responses for the self-consistency arm were collected under a cache key that
includes the sampling temperature; an earlier key omitted it and could have
replayed a temperature-$0$ response, so that arm was collected afresh.

\subsection{No detectable self-favoritism at this resolution}

A model might go easy on trajectories it produced. Under the neutral prompt each
auditor audits both producers' task sets ($n = 120$ per cell), and neither is more
lenient on itself: deepseek-v4-pro scores $0.700$ on its own trajectories against
$0.667$ on the other's (gap $-0.033$, Fisher $p = 0.68$), and glm-5 scores $0.583$
on both (gap exactly $0.000$). Two limits keep this from being strong evidence of
absence. The design's minimum detectable effect is roughly $0.17$ absolute recall
at this sample size, so only a large self-preference would have shown. And with
two producers each auditor's ``cross'' condition is a single other model, which
aliases producer difficulty with self-favoritism perfectly.

\section{Discussion}\label{sec:synthesis}

The two studies answer the question in the introduction with different
controls: Study~A holds the response tape fixed before an execution contrast
counts as mechanical, and Study~B adds matched defect conditions and
specificity outcomes before a recall score counts as audit quality.

\paragraph{What the headline summaries leave out.}
In Study~A, all $89$ agreeing paths come
from the nearly inactive model; pooled agreement should not be read as stability
of active policies. In Study~B, coverage of the target labels can coexist with
many unsupported findings. A descriptive decomposition of the existing
dual-defect responses shows that deepseek-v4-pro reports both true labels in
$78/100$ tasks, but adds extra findings in $57$ of those cases, leaving $21/100$
strictly correct sets. The corresponding counts for glm-5 are $23$, $16$, and
$7$. None of these both-target cases is a parse failure. This explains the gap
between coverage and exact correctness.

\begin{figure}[t]
\centering
\includegraphics[width=\textwidth]{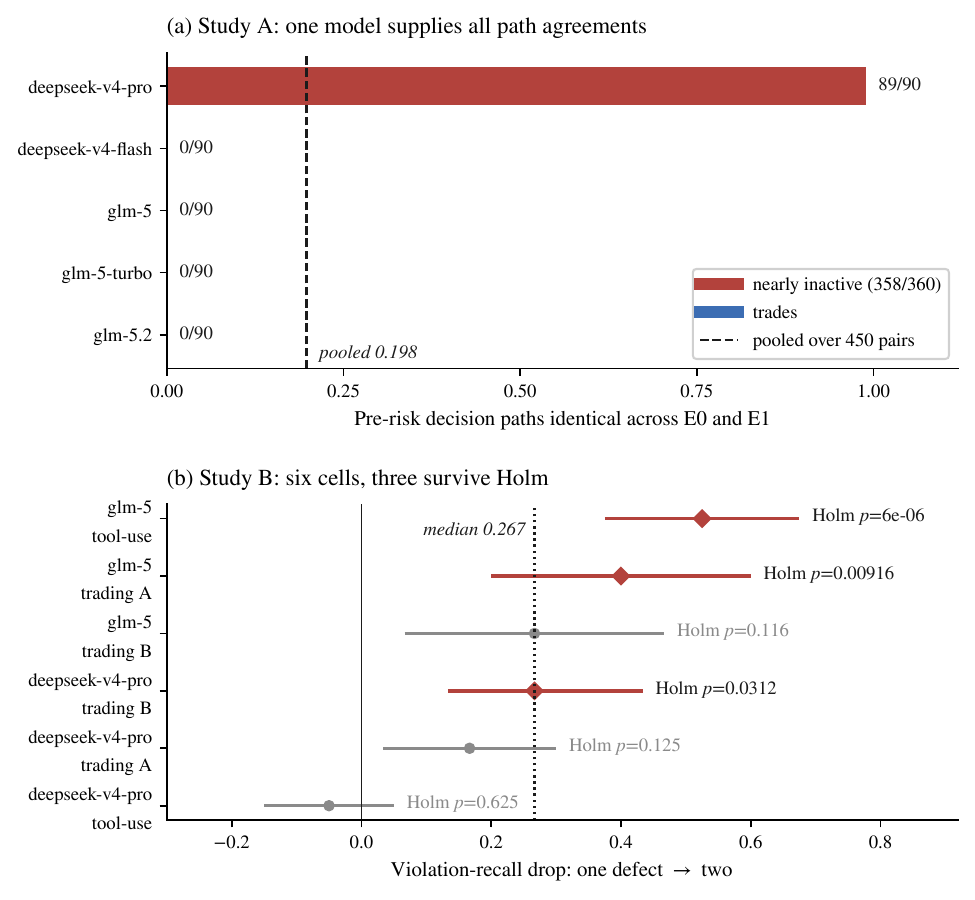}
\caption{Two aggregate summaries and their cell structure. (a) Parsed-path
agreement in the independently sampled E0/E1 source runs. The pooled $19.8\%$
consists of $89/90$ agreements from one nearly inactive model and $0/90$ from
each active model. This shows that the diagonal is not a fixed-output comparison;
it does not show that execution caused the divergence, since most divergent
pairs already differ in the first response. (b) The six cells recorded in the
internally hash-bound multi-label plan. Five target-recall drops are positive,
three survive Holm correction, and one is negative.}
\label{fig:pooling}
\end{figure}

Figure~\ref{fig:pooling} shows why pooled summaries are insufficient here. The
model-level path split changes the interpretation of $19.8\%$, and the six audit
cells show both heterogeneity and an exception to the median direction.

\paragraph{Reporting checks.}
Four checks follow from the two studies. State the quantity the design
estimates. State what is compared and what is held fixed: the response tape in
Study~A, the source case and audit instruction across defect conditions in
Study~B. Report what the headline score leaves out: activity and ranking
uncertainty beside an execution contrast, unsupported findings and zero-defect
behaviour beside recall. Keep the conclusion within the design's resolution: a
rank correlation of $1.000$ over ten seeds coexists with exact-order agreement
in only $0.26$ of bootstrap resamples, and detecting both targets often
coexists with an incorrect full answer.

\section{Limitations}\label{sec:limits}

Study~A rests on ten shared seed clusters, three synthetic settings that occupy
the same 24-day upward phase, one uncalibrated stress parameterization, and a
provider mix that includes an almost inactive model. How an adaptive agent
would respond to E1 fills, and how the contrast behaves over a downward or
sideways phase, are outside the design.

Study~B uses two hosted API snapshots, synthetic substrates, and one coupled
violation-plus-edit construction. Other defect combinations, models, prompts,
and deployment distributions are untested.

\paragraph{Scope.} The substrate is financial, the conventions are not. We make
no claim that any model studied here is better or worse at trading, and the
returns reported are properties of a simulator under stated assumptions.

\section{Conclusion}\label{sec:conclusion}

In Study~A, holding the response tape fixed turns an ambiguous comparison between
E0 and E1 into a mechanical contrast of $-0.0170$ in total return, while ten
seed clusters leave the model ranking open. In Study~B, recall measured with
one defect does not carry over to the same case with two, and the auditor that
most often includes both targets also has low precision, frequent zero-defect
false positives, and only $21/100$ exact dual-defect sets.

A performance score supports only the claim that its controls identify and its
scoring covers. Stating that estimand, reporting activity and ranking
uncertainty beside an execution contrast, and reporting specificity beside
recall keep a limited statistic from carrying a broader claim than the design
supports.

\section*{Reproducibility}
\addcontentsline{toc}{section}{Reproducibility}

The accompanying research artifact contains analysis code, figure sources,
frozen inputs, and a strict offline verifier. It is released at
\url{https://github.com/weich97/llm-evaluation-boundaries-artifact}.
For Study~A it supports statistical
reanalysis of the $1{,}800$ recorded replay outcomes, not regeneration of the
underlying provider calls or execution trajectories. For Study~B's primary
experiment it includes $300$ tasks, their answer keys, $600$ structured results,
and the original response text, including all nine parse failures. Prompts are
reconstructed from the task artifacts and checked against recorded request
hashes. The verifier reruns the parser, task-rule checks, scoring, and analysis.
Legacy re-scoring, interventions, and self-audit summaries are reproducible from
released structured findings and keys; raw responses for those secondary
collections are not included.

The plan, manifests, and task trees use content digests to detect disagreement.
Because they remain author-writable and have no third-party timestamp, those
digests do not prove that the plan predates collection. A separate
outcome-dependent rule, outside the hashed analysis plan, guided whether to
pursue Study~B as a standalone paper, merge it into a broader report, or release
it as a negative result.

The entry point is \texttt{scripts/verify\_tmlr\_supplement.py}. It checks the
file manifest and complete required inputs, rejects missing or duplicate result
keys, recomputes tables, and regenerates the appendix and figures. Required
checks cannot be skipped. A separate checker validates selected printed values
against the manuscript source when that source is present; the repository
ships code and data, not the manuscript. Neither verifier establishes provider
authenticity or corpus representativeness. The included coverage statement
identifies the reconstruction level for each study and secondary analysis; no
new provider call is needed.

\section*{Use of Generative AI}
\addcontentsline{toc}{section}{Use of Generative AI}

The author planned and led the study, reviewed the literature, developed the
methods, designed the experiments, analyzed the results, and wrote the paper.
Under the author's guidance, generative AI tools helped write code, gave
feedback on the experimental design, and helped check the mathematical and
statistical analyses. The tools also helped edit the text, write plotting
scripts, prepare the figures, and prepare and check the data and code package
for release with the paper.

\bibliography{refs}

\appendix
\section*{Appendix: per-cell tables}
\addcontentsline{toc}{section}{Appendix: per-cell tables}

The appendix reports the per-cell values underlying the summaries in the main
text. All tables are generated directly from the released CSV files by
\texttt{scripts/build\_appendix\_tables.py}.

\makeatletter
\setlength{\@fptop}{0pt}
\makeatother
\input{appendix_tables}

\end{document}

%% file: appendix_tables.tex

\begin{table}[htbp]
\centering
\caption{Study A, every estimand of the fixed-tape $2\times2$ replay, on the four outcomes the body draws on. Intervals are seed-cluster bootstrap ($10{,}000$ draws, $10$ clusters, $450$ matched pairs). The replay records five further outcomes; they are in \texttt{factorial\_estimands.csv}.}
\label{tab:app-factorial}
\small
\begin{tabular}{llrr}
\toprule
Outcome & Estimand & Estimate & 95\% CI \\
\midrule
total return & $D_{\mathrm{E0}}$: execution $\mid$ tape from E0 & -0.0193 & $[-0.0273, -0.0121]$ \\
 & $D_{\mathrm{E1}}$: execution $\mid$ tape from E1 & -0.0147 & $[-0.0197, -0.0100]$ \\
 & $\bar D$: balanced destination & -0.0170 & $[-0.0230, -0.0117]$ \\
 & $O_{\mathrm{E0}}$: response origin $\mid$ replay at E0 & +0.0012 & $[-0.0075, +0.0101]$ \\
 & $O_{\mathrm{E1}}$: response origin $\mid$ replay at E1 & +0.0059 & $[-0.0053, +0.0176]$ \\
 & $\bar O$: balanced tape origin & +0.0035 & $[-0.0061, +0.0136]$ \\
 & $D_{\mathrm{E1}}-D_{\mathrm{E0}}$: interaction & +0.0047 & $[-0.0005, +0.0105]$ \\
 & $\bar D+\bar O$: observed diagonal & -0.0135 & $[-0.0238, -0.0025]$ \\
\midrule
Sharpe & $D_{\mathrm{E0}}$: execution $\mid$ tape from E0 & -3.3944 & $[-3.8297, -3.0142]$ \\
 & $D_{\mathrm{E1}}$: execution $\mid$ tape from E1 & -1.9269 & $[-2.4712, -1.3757]$ \\
 & $\bar D$: balanced destination & -2.6607 & $[-3.1269, -2.2312]$ \\
 & $O_{\mathrm{E0}}$: response origin $\mid$ replay at E0 & +0.0796 & $[-0.3052, +0.4790]$ \\
 & $O_{\mathrm{E1}}$: response origin $\mid$ replay at E1 & +1.5470 & $[+1.1073, +2.0632]$ \\
 & $\bar O$: balanced tape origin & +0.8133 & $[+0.4409, +1.2386]$ \\
 & $D_{\mathrm{E1}}-D_{\mathrm{E0}}$: interaction & +1.4674 & $[+1.1168, +1.8250]$ \\
 & $\bar D+\bar O$: observed diagonal & -1.8473 & $[-2.4076, -1.1905]$ \\
\midrule
max drawdown & $D_{\mathrm{E0}}$: execution $\mid$ tape from E0 & -0.0050 & $[-0.0070, -0.0031]$ \\
 & $D_{\mathrm{E1}}$: execution $\mid$ tape from E1 & -0.0035 & $[-0.0062, -0.0013]$ \\
 & $\bar D$: balanced destination & -0.0042 & $[-0.0065, -0.0025]$ \\
 & $O_{\mathrm{E0}}$: response origin $\mid$ replay at E0 & +0.0004 & $[-0.0006, +0.0019]$ \\
 & $O_{\mathrm{E1}}$: response origin $\mid$ replay at E1 & +0.0020 & $[+0.0004, +0.0033]$ \\
 & $\bar O$: balanced tape origin & +0.0012 & $[+0.0003, +0.0023]$ \\
 & $D_{\mathrm{E1}}-D_{\mathrm{E0}}$: interaction & +0.0015 & $[-0.0004, +0.0033]$ \\
 & $\bar D+\bar O$: observed diagonal & -0.0030 & $[-0.0047, -0.0014]$ \\
\midrule
fill rate & $D_{\mathrm{E0}}$: execution $\mid$ tape from E0 & -0.0682 & $[-0.0718, -0.0642]$ \\
 & $D_{\mathrm{E1}}$: execution $\mid$ tape from E1 & -0.0496 & $[-0.0519, -0.0473]$ \\
 & $\bar D$: balanced destination & -0.0589 & $[-0.0613, -0.0563]$ \\
 & $O_{\mathrm{E0}}$: response origin $\mid$ replay at E0 & -0.0022 & $[-0.0067, +0.0000]$ \\
 & $O_{\mathrm{E1}}$: response origin $\mid$ replay at E1 & +0.0164 & $[+0.0098, +0.0219]$ \\
 & $\bar O$: balanced tape origin & +0.0071 & $[+0.0017, +0.0109]$ \\
 & $D_{\mathrm{E1}}-D_{\mathrm{E0}}$: interaction & +0.0186 & $[+0.0147, +0.0223]$ \\
 & $\bar D+\bar O$: observed diagonal & -0.0518 & $[-0.0578, -0.0474]$ \\
\bottomrule
\end{tabular}
\end{table}

\begin{table}[t]
\centering
\caption{Study B, the six auditor $\times$ source cells recorded in the internally hash-bound multi-label plan. Recall is on the target violation, within matched triplets; the discordant counts are the McNemar pairs. Intervals are unadjusted 95\% paired percentile-bootstrap confidence intervals based on $10{,}000$ resamples. $p$-values are from two-sided exact McNemar tests with Holm correction across the six primary comparisons.}
\label{tab:app-cells}
\scriptsize
\begin{tabular}{llrrrrrrr}
\toprule
Auditor & Source & $n$ & Single & Dual & Drop & 95\% CI & $b/c$ & Holm $p$ \\
\midrule
deepseek-v4-pro & tool-use & 40 & 0.900 & 0.950 & -0.050 & $[-0.150, +0.050]$ & 1/3 & 0.625 \\
deepseek-v4-pro & trading A & 30 & 0.900 & 0.733 & +0.167 & $[+0.033, +0.300]$ & 5/0 & 0.125 \\
\bf deepseek-v4-pro & \bf trading B & 30 & 0.967 & 0.700 & \bf +0.267 & $[+0.133, +0.433]$ & 8/0 & \bf 0.0312 \\
\bf glm-5 & \bf tool-use & 40 & 1.000 & 0.475 & \bf +0.525 & $[+0.375, +0.675]$ & 21/0 & \bf 5.7e-06 \\
\bf glm-5 & \bf trading A & 30 & 0.833 & 0.433 & \bf +0.400 & $[+0.200, +0.600]$ & 13/1 & \bf 0.00916 \\
glm-5 & trading B & 30 & 0.433 & 0.167 & +0.267 & $[+0.067, +0.467]$ & 10/2 & 0.116 \\
\bottomrule
\end{tabular}
\end{table}